%% file: main.tex
\documentclass[10pt,twocolumn,letterpaper]{article}

\usepackage{cvpr}              
\usepackage{multirow}
\usepackage{algorithm}
\usepackage{algorithmic}
\usepackage{amsfonts,amssymb}
\usepackage{booktabs}
\usepackage{comment}

\usepackage{colortbl}
\usepackage{color}

\usepackage{bibentry}
\usepackage{multirow}
\usepackage{fancyhdr}
\usepackage{amsmath}
\usepackage{verbatim}
\usepackage{amsthm}
\usepackage{tabularx}
\usepackage{bbding}

\newtheorem{theorem}{Theorem}
\newtheorem{corollary}[theorem]{Corollary}
 
\usepackage{newfloat}
\usepackage{listings}
\usepackage{graphicx}


\definecolor{cvprblue}{rgb}{0.21,0.49,0.74}
\usepackage[pagebackref,breaklinks,colorlinks,allcolors=cvprblue]{hyperref}
\usepackage[capitalize]{cleveref}
\Crefname{section}{Section}{Sections}
\Crefname{table}{Table}{Tables}
\crefname{figure}{Figure}{Figures}
\crefname{equation}{Equation}{Equations}
\def\confName{CVPR}
\def\confYear{2025}

\title{Night-to-Day Translation via Illumination Degradation Disentanglement}

\author{
   Guanzhou Lan\textsuperscript{\rm 1,2},
    Yuqi Yang \textsuperscript{\rm 1},
    Zhigang Wang \textsuperscript{\rm 1,2}, \\
    Dong Wang \textsuperscript{\rm 2},
    Bin Zhao \textsuperscript{\rm 1,2 $\dagger$},
    Xuelong Li \textsuperscript{\rm 1 $\dagger$}\\\\
    \textsuperscript{\rm 1} Northwestern Polytechnical University,
    \textsuperscript{\rm 2} Shanghai AI Laboratory\\
}

\begin{document}
\maketitle
\input{sec/0_abstract}    
\input{sec/1_intro}

\input{sec/2_related_work}

\input{sec/3_method}
\input{sec/4_exp}

\input{sec/5_conclusion}

{
    \small
    \bibliographystyle{ieeenat_fullname}
    \bibliography{main}
}


\end{document}

%% file: sec/0_abstract.tex
\begin{abstract}
Night-to-Day translation (Night2Day) aims to achieve day-like vision for nighttime scenes. However, processing night images with complex degradations remains a significant challenge under unpaired conditions. Previous methods that uniformly mitigate these degradations have proven inadequate in simultaneously restoring daytime domain information and preserving underlying semantics. In this paper, we propose \textbf{N2D3} (\textbf{N}ight-to-\textbf{D}ay via \textbf{D}egradation \textbf{D}isentanglement) to identify different degradation patterns in nighttime images. Specifically, our method comprises a degradation disentanglement module and a degradation-aware contrastive learning module. Firstly, we extract physical priors from a photometric model based on Kubelka-Munk theory. Then, guided by these physical priors, we design a disentanglement module to discriminate among different illumination degradation regions. Finally, we introduce the degradation-aware contrastive learning strategy to preserve semantic consistency across distinct degradation regions. Our method is evaluated on two public datasets, demonstrating a significant improvement in visual quality and considerable potential for benefiting downstream tasks.
\makeatletter{\renewcommand*{\@makefnmark}{}
\footnotetext{\textsuperscript{$\dagger$}Corresponding authors.
}}

\end{abstract}

%% file: sec/1_intro.tex
\section{Introduction}

Nighttime images often suffer from severe information loss, posing significant challenges to both human visual recognition and computer vision tasks including detection, segmentation, \emph{etc.} \cite{kennerley20232pcnet}. In contrast, daytime images exhibit rich content and intricate details. Achieving day-like nighttime vision remains a primary objective in nighttime perception, sparking numerous pioneering works \cite{Yu_Infrared-to-Visible_2023_CVPR}.
\begin{figure}[t]
\centering
\includegraphics[width=0.99\linewidth]{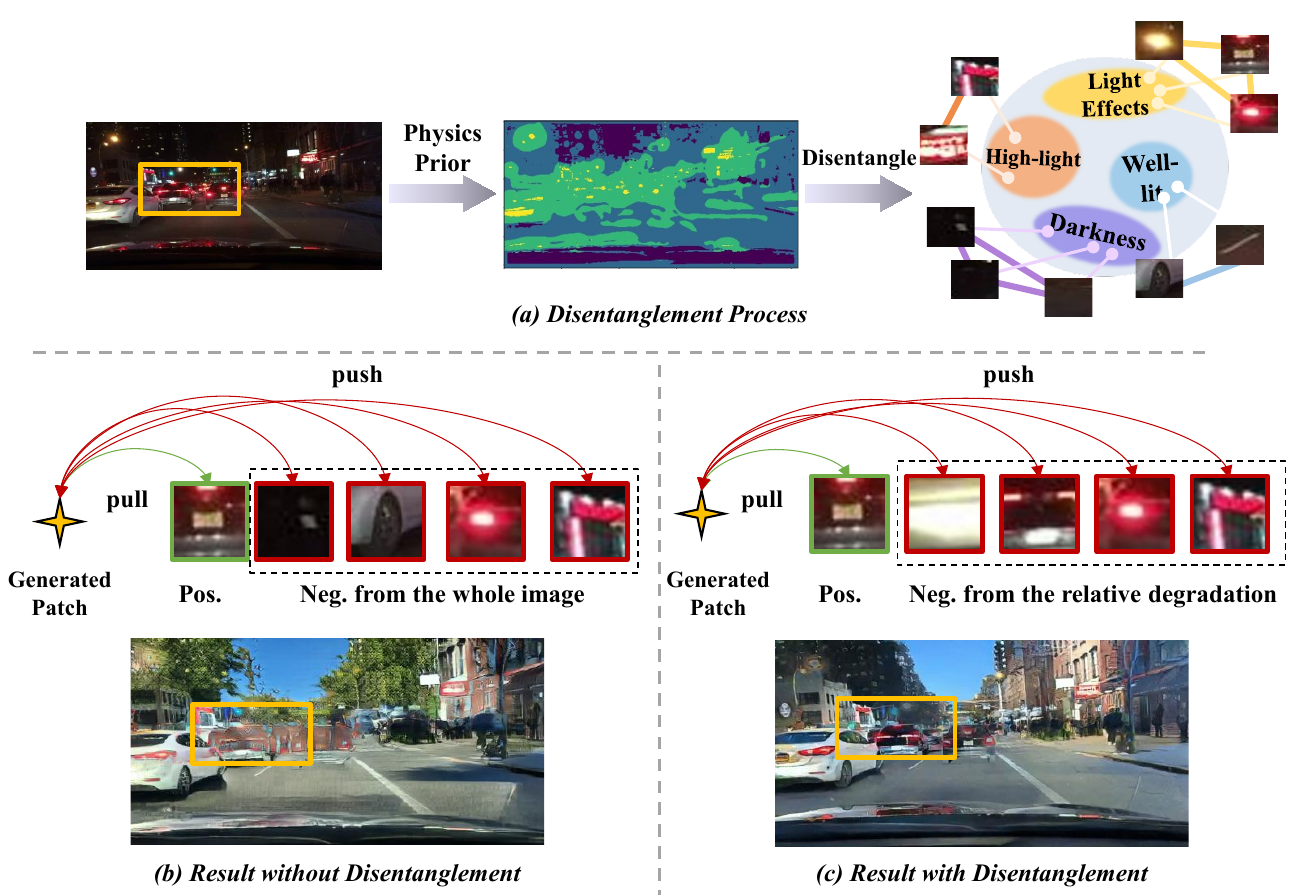} 
\vspace{-0.3cm}
\caption{Illustration of our motivation. (a) The disentanglement process leverages physical priors. (b) Vanilla structure regularization and the corresponding results. (c) The proposed disentangled regularization and the corresponding results. }
\vspace{-0.6cm}
\label{fig:top}
\end{figure}
Night-to-Day image translation (Night2Day) offers a comprehensive solution to achieve day-like vision at night. The primary goal is to transform images from nighttime to daytime while maintaining their underlying semantic structure. However, achieving this goal is challenging. Since no ground truth is available for daytime images, we must capture the underlying semantic structure within the complex, degraded nighttime images, which presents additional challenges compared to other image translation tasks.

Early efforts in Night2Day primarily focused on addressing these challenges by applying image translation techniques such as cycle-consistent learning and domain-invariant learning \cite{anoosheh2019night, zheng2020forkgan}. These methods introduced advanced generative adversarial models to the field but overlooked the core challenge of representing the underlying semantic structure, which led to additional artifacts during translation. AUGAN was among the first to recognize the importance of uncovering the underlying structure, incorporating uncertainty to better preserve the original structure \cite{kwak2021adverse}. Additionally, some approaches have used daytime images with nearby GPS locations to aid in coarse structure regularization \cite{xia2023image}. While these methods attempt to leverage statistical priors to uncover the underlying structure, they often neglect the complex degradations that occur at nighttime, applying structure regularization uniformly and resulting in severe artifacts. More recent approaches have adopted auxiliary human annotations, such as segmentation maps and bounding boxes, to maintain semantic consistency \cite{kim2022instaformer, Song_Lee_Seong_Min_Kim_2023}. Despite their potential, these methods are labor-intensive and difficult to implement, particularly for nighttime scenes that are beyond human cognition.

Previous research has sought to capture the underlying semantic structure using statistical priors. However, these approaches often produce suboptimal results, as the domain-invariant features learned from such priors lack physical significance. This raises the following question: \textit{Could a physical prior offer a more efficient way for Night2Day to extract the underlying semantic structure?} To address this, we first define the domain-invariant features of the nighttime domain from a physical perspective. Specifically, the reflectance under equal energy but uneven illumination corresponds to the domain-invariant regions that we refer to as \textit{well-lit}. These well-lit regions typically exhibit intermediate illumination intensities in nighttime images, as they are illuminated under normal conditions without producing strong reflections.
However, a key observation is that colored illumination in the nighttime represents a domain-specific feature, yet it shares similar intensity levels with other well-lit regions. Treating the effects of light and other well-lit regions equally may negatively impact the results, as demonstrated in \cref{fig:top}.

Based on these observations, we propose N2D3 (\textbf{N}ight to \textbf{D}ay via \textbf{D}egradation \textbf{D}isentanglement), which utilizes Generative Adversarial Networks (GANs) to bridge the domain gap between nighttime and daytime in a degradation-aware manner, as illustrated in \cref{fig:main}. N2D3 consists of two key modules: physics-informed degradation disentanglement and degradation-aware contrastive learning, both of which preserve the semantic structure of nighttime images. In the disentanglement of nighttime degradation, a photometric model tailored to nighttime scenes is conducted to extract physical priors. Subsequently,we propose a disentanglement strategy to separate the distinct patterns observed in nighttime images. Since the intensity of illumination is the most important criterion at nighttime, we begin by categorizing nighttime images into three non-overlapping regions: high-light, well-lit, and darkness. Moreover, to disentangle light effects from well-lit regions, we demonstrate both theoretically and empirically that a color-invariance property can effectively isolate light effects from well-lit regions.

Building on this, degradation-aware contrastive learning is designed to constrain the similarity of the source and generated images in different regions. It comprises disentanglement-guided sampling and reweighting strategies. The sampling strategy mines valuable anchors and hard negative examples, while the reweighting process assigns their weights. They enhance vanilla contrastive learning by prioritizing valuable patches with appropriate attention. Ultimately, our method yields highly faithful results that are visually pleasing and beneficial for downstream vision tasks including keypoint matching and semantic segmentation.
Our contributions are summarized as follows:

(1) We propose the N2D3 translation method based on the illumination degradation disentanglement module, which enables degradation-aware restoration of nighttime images.

(2) We present a novel degradation-aware contrastive learning module to preserve the semantic structure of generated results. The core design incorporates disentanglement-guided sampling and reweighting strategies, which greatly enhance the performance of vanilla contrastive learning.

(3) Experimental results on two public datasets underscore the significance of considering distinct degradation types in nighttime scenes. Our method achieves state-of-the-art performance in visual effects and downstream tasks.

%% file: sec/2_related_work.tex
\section{Related Work}

\textbf{Unpaired Image-to-Image Translation.}  Unpaired image-to-image translation addresses the challenge of lacking paired data, providing an effective self-supervised learning strategy. To overcome the efficiency limitations of traditional cycle-consistency learning, Park \emph{et al.},  first introduces contrastive learning to this domain, achieving efficient one-sided learning\cite{park2020contrastive}. Following this work, several studies have improved the contrastive learning by generating hard negative examples \cite{wang2021instance}, re-weighting positive-negative pairs \cite{Zhan_2022_CVPR}, and selecting key samples \cite{hu2022qs}. Furthermore, other constraints, such as density \cite{decent} and path length \cite{santa}, have been explored in unpaired image translation. However, all these works neglect physical priors in the nighttime, leading to suboptimal results in Night2Day.

\begin{figure*}[ht]
\centering
\includegraphics[width=0.9\linewidth]{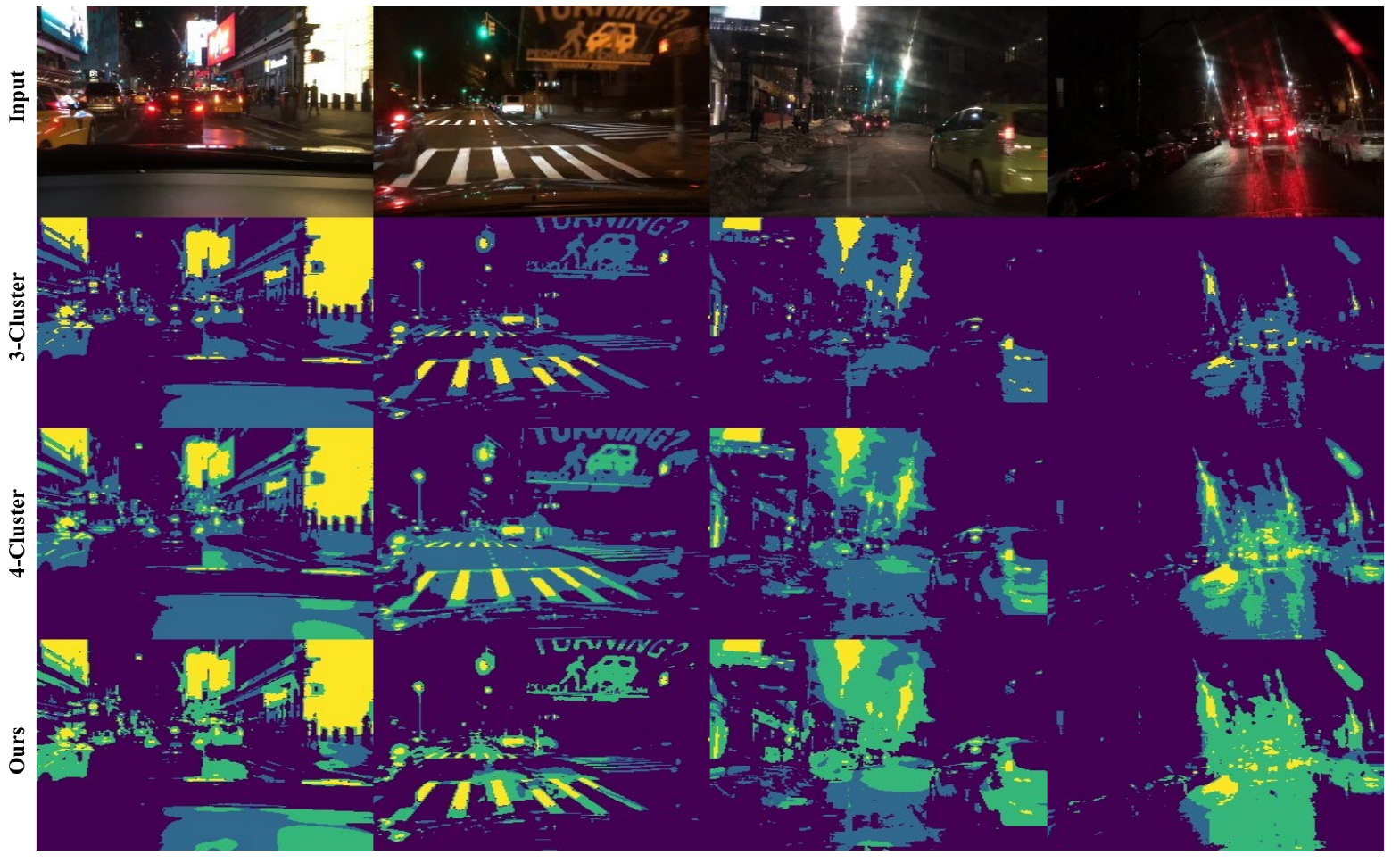} 
\vspace{-0.2cm}
\caption{The first row displays nighttime images, while the second row shows the disentanglement results using 3-Cluster K-means. The third row presents the disentanglement results with 4-Cluster K-means, and the last row shows the results from our physics-informed disentanglement. The color progression from \textbf{\textcolor{blue}{blue}}, \textbf{\textcolor{blue!50}{light blue}}, \textbf{\textcolor{green!70}{green}} to \textbf{\textcolor{yellow!100}{yellow}} corresponds to the following regions: darkness, well-lit, light effects, and high-light, respectively.}
\vspace{-0.4cm}
\label{fig:dis}
\end{figure*}

\noindent \textbf{Nighttime Domain Translation.}
Domain translation techniques have been applied to address adverse nighttime conditions. An early contribution is made by Anoosheh \emph{et al.}, which demonstrates the effectiveness of cycle-consistent learning in Night2Day\cite{anoosheh2019night}. Following this, many works incorporate different modules into cycle-consistent learning to enhance structural modeling capabilities. Zheng \emph{et al.} incorporate a fork-shaped encoder to enhance visual perceptual quality\cite{zheng2020forkgan}. AUGAN  employs uncertainty estimation to mine useful features in nighttime images\cite{kwak2021adverse}.  Fan \emph{et al.} explore inter-frequency relation knowledge to streamline the Night2Day process\cite{fan2023learning}. Xia \emph{et al.} utilize nearby GPS locations to form paired night and daytime images, providing weak supervision\cite{xia2023image}. Some other studies incorporate human annotations to impose structural constraints, overlooking the practical difficulty of acquiring such annotations at nighttime with multiple degradations \cite{jeong2021memory, kim2022instaformer, Song_Lee_Seong_Min_Kim_2023}. 
To address the concerns of the aforementioned methods, the proposed N2D3 explores patch-wise contrastive learning with physical guidance, so as to achieve degradation-aware Night2Day. N2D3 is free of human annotations and offers comprehensive structural modeling to provide faithful translation results.

%% file: sec/3_method.tex
\section{Methods}\label{sec: Methods}
\begin{figure*}[t]
\centering
\includegraphics[width=0.99\linewidth]{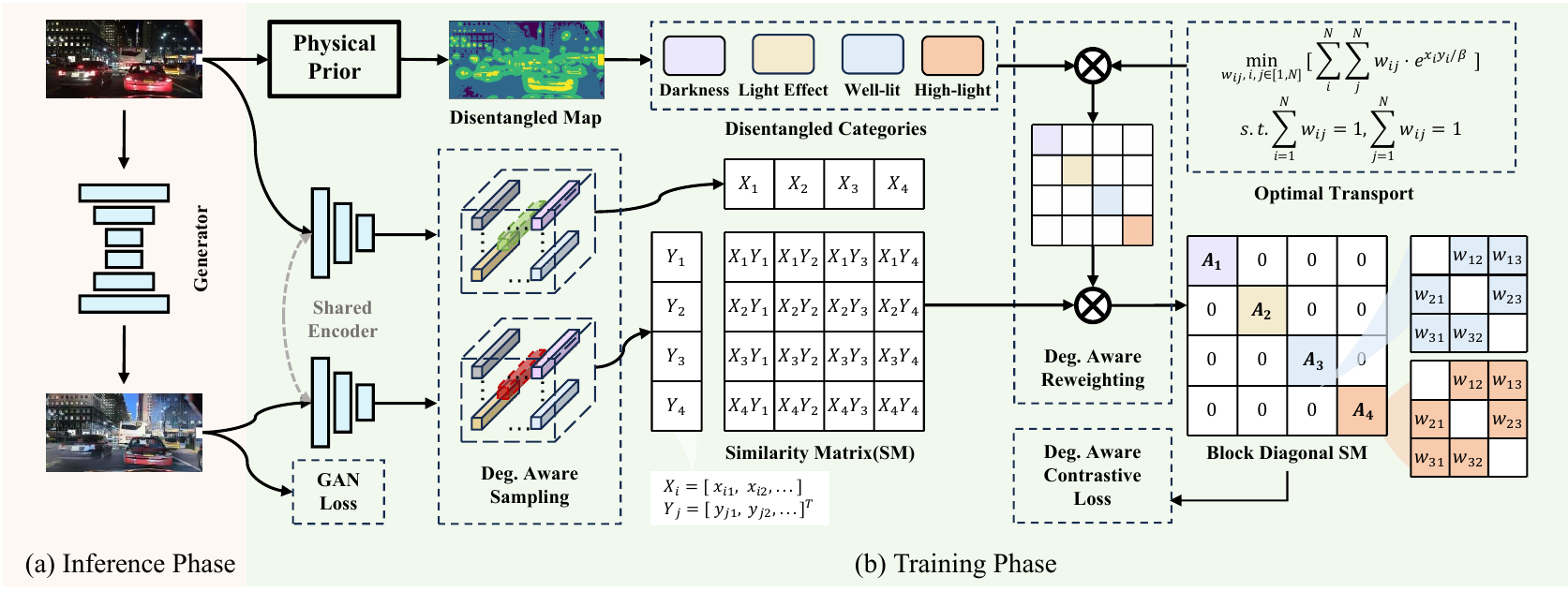} 
\vspace{-0.3cm}
\caption{ The overall architecture of the proposed N2D3 method. The training phase contains the physical prior informed degradation disentanglement module and degradation-aware contrastive learning module. They are utilized to optimize the ResNet-based generator which is the main part in the inference phase.}
\vspace{-0.5cm}
\label{fig:main}
\end{figure*}

Given nighttime image $\bf{I}_\mathcal{N} \in \mathcal{N}$ and daytime image $\bf{I}_\mathcal{D} \in \mathcal{D}$, the goal of Night2Day is to translate images from nighttime to daytime while preserving content semantic consistency. This involves the construction of a mapping function $\mathcal{F}$ with parameters $\theta$, which can be formulated as $\mathcal{F}_\theta: \mathbf{I}_\mathcal{N} \rightarrow \mathbf{I}_\mathcal{D}$. Our method N2D3 is illustrated in \cref{fig:main}. To train a generator for Night2Day, we employ GANs as the overall learning framework to bridge the domain gap between nighttime and daytime. Our core design, consisting of the degradation disentanglement module and the degradation-aware contrastive learning module, aims to preserve the structure from the source images and suppress artifacts.

In this section, we first introduce physical priors in the nighttime environment, and then describe the degradation disentanglement module and the degradation-aware contrastive learning module, respectively.

\subsection{Physical Priors for Nighttime Environment}\label{sec: Physical}
The illumination degradations\cite{geusebroek2001color} at night are primarily categorized as four types: darkness, well-lit regions, high-light regions, and light effects. As shown in Figure \ref{fig:dis}, well-lit represents the diffused reflectance under normal light, while the light effects denote phenomena such as flare, glow, and specular reflections. Intuitively, these regions can be disentangled through the analysis of illumination distribution. Among these degradation types, darkness and high-light are directly correlated with illuminance and can be effectively disentangled through illumination estimation.

As a common practice, we estimate the illuminance map $L$ by utilizing the maximum RGB channel of image $\mathbf{I}_\mathcal{N}$ as $L = \max_{c \in R, G, B} \mathbf{I}^c_\mathcal{N}$. Then k-means is employed to acquire three clusters representing darkness, well-lit, and high-light regions. These clusters are aggregated as masks $M_d$, $M_n$, $M_h$.  However, the challenge arises with light effects that are mainly related to the illumination. Light effects regions tend to intertwine with well-lit regions when using only the illumination map, as they often share similar illumination densities. To disentangle light effects from well-lit regions, we need to introduce additional physical priors.

To extract the physical priors for disentangling light effects, we develop a photometric model derived from Kubelka-Munk theory \cite{geusebroek2001color}. This model characterizes the spectrum of light $E$ reflected from an object as follows:
\begin{equation} 
    \begin{aligned}
        \ & E(\lambda, x) = e(\lambda, x)(1-\rho_f(x))^2R_{\infty}(\lambda, x) + e(\lambda, x)\rho_f(x),
    \end{aligned}	
    \label{eq: K-M}
\end{equation}
here $x$ represents the horizontal component for analysis, while the analysis of the vertical component $y$ is the same as the horizontal component. $\lambda$ corresponds to the wavelength of light. $e(\lambda,x)$ signifies the spectrum, representing the illumination density and color. $\rho_f$ stands for the Fresnel reflectance coefficient. 

We assume that materials are uniform and homogeneous within a local area under normal conditions. Specifically, the optical properties of a material within a small region are described by the function $R(\lambda)$, which characterizes the material's properties as a function of wavelength and is independent of location. Similar assumption is also used in materials science, optics and computer vision. Under this assumption, we can simplify the reflectivity function 
$R\infty(\lambda ,x)$ in the local area to $cR(\lambda )$
, where $c$ is coefficient that describe the material type. To model the global nighttime conditions, we introduce the material spatial distribution function $C(x)$ is defined as:  $C:\mathbb R \rightarrow c_1,c_2,\dots,c_m$. With $C(x)$, we can model more complex nighttime scenes with diverse material types at macro scales as:
\begin{equation}
    \begin{aligned}
        \ & R_\infty(\lambda, x) =R(\lambda)C(x) .
    \end{aligned}	
    \label{eq: R_decom}
\end{equation}  

Since the mixture of light effects and well-lit regions has been obtained previously, the core of disentangling light effects from well-lit regions lies in separating the illumination $e(\lambda, x)$ and reflectance components $R(\lambda)C(x)$. Note that the Fresnel reflectance coefficient $\rho_f(x)$ approaches 0 in reflectance-dominating well-lit regions, while $\rho_f(x)$ approaches 1 in illumination-dominating light effects regions. According to \cref{eq: K-M}, the photometric model for the mixture of light effects and well-lit is formulated as:
\begin{equation}
    \begin{aligned}
            E(\lambda, x) =\begin{cases}
          e(\lambda, x), & \text{if } x \notin \Omega \\
           e(\lambda, x)R(\lambda)C(x), & \text{if } x \in \Omega \\
        \end{cases},
    \end{aligned}	
    \label{eq: K-M-our}
\end{equation}
where $\Omega$ denotes the reflectance-dominating well-lit regions. Subsequently, we observe that the following color invariant response to the regions with high color saturation, which is suitable to extract the illumination, as outlined in the Corollary~\ref{cor:N}.

\begin{corollary}[Proof in the supplementary material]
\label{cor:N}
Under the assumption of local uniformity and homogeneity, a complete and irreducible set of invariants for the color illumination spectrum is given by:
\begin{equation}
    \begin{aligned}
        N_{{\lambda}^m x^n} &= \frac{\partial^{m+n-2}}{\partial \lambda^{m-1}  \partial x^{n-1}} \frac{\partial}{\partial x} \Bigg\{\frac{1}{E(\lambda, x)} \frac{\partial E(\lambda, x)}{\partial \lambda} \Bigg\} \\   
        &= \frac{\partial^{m+n-1}}{\partial \lambda^{m-1}  \partial x^n} \Bigg\{\frac{1}{e(\lambda, x)} \frac{\partial e(\lambda, x)}{\partial \lambda} \Bigg\}.
    \end{aligned}	
    \label{eq: N-2}
\end{equation}
\end{corollary}

Corollary~\ref{cor:N} demonstrate that the invariant $N_{{\lambda}^m x^n}$ captures the features only related to illumination $e(\lambda, x)$. Consequently, we assert that $N_{{\lambda}^m x^n}$ functions as a light effects detector because light effects are mainly related to the illumination. It allows us to design the illumination disentanglement module based on this physical prior.

\subsection{Degradation Disentanglement Module} \label{sec: dis}
In this subsection, we will elucidate how to incorporate the invariant for extracting light effects into the disentanglement in computation. As common practice, the following second and third-order components, both horizontally and vertically, are taken into account in the practical calculation of the final invariant, which is denoted as $N$:
\begin{equation} 
    \begin{aligned}
        &N= \sqrt{N_{\lambda x}^2 + N_{\lambda \lambda x}^2 + N_{\lambda y}^2 + N_{\lambda \lambda y}^2}.
    \end{aligned}	
    \label{eq: N-4}
\end{equation}
here $N_{\lambda x}$ and $N_{\lambda \lambda x}$ can be computed through $E(\lambda, x)$ by simplifying \cref{eq: N-2}. The calculation of $N_{\lambda y}$ and $N_{\lambda \lambda y}$ are the same. Specifically,

\begin{equation} 
    \begin{aligned}
       N_{\lambda x} &=\frac{E_ {\lambda x}E-E_\lambda E_x}{E^2}, \\  
       N_{\lambda \lambda x} &= \frac{E_ {\lambda \lambda x}E^2-E_{\lambda \lambda} E_xE - 2E_{\lambda x}E_\lambda E + 2 E_\lambda^2E_x}{E^3} ,   \\
    \end{aligned}
    \label{eq: N-3}
\end{equation}
where $E_x$ and $E_\lambda$ denote the partial derivatives of $x$ and $\lambda$. 

To compute each component in the invariant $N$, we develop a computation scheme starting with the estimation of $E$ and its partial derivatives $E_\lambda$ and $E_{\lambda \lambda}$ using the Gaussian color model:
\begin{equation} 
    \begin{aligned}
        \begin{bmatrix}
            E(x,y) \\
            E_\lambda(x,y)  \\
            E_{\lambda \lambda}(x,y)
        \end{bmatrix} = 
        \begin{bmatrix}
            &0.06,&0.63, &0.27 \\
            &0.3, &0.04, &-0.35 \\
            &0.34, &-0.6, &0.17
        \end{bmatrix}
        \begin{bmatrix}
            R(x,y) \\
            G(x,y) \\
            B(x,y) \\
        \end{bmatrix},
    \end{aligned}	
    \label{eq: Guassian}
\end{equation}
where $x,y$ are pixel locations of the image. Then, the spatial derivatives $E_x$ and $E_y$ are calculated by convolving $E$ with Gaussian derivative kernel $g$ and standard deviation $\sigma$:
\begin{equation}
    \begin{aligned}
        \ & E_x(x, y, \sigma) = \sum_{t \in \mathbf{Z} }{E(t,y)\frac{\partial g(x-t, \sigma)}{\partial x}},
    \end{aligned}	
    \label{eq: E_x}
\end{equation}
where $t$ denotes the index of the horizontal component $x$ and $\mathbf{Z}$ represents set of integers. The spatial derivatives for $E_{\lambda x}$ and $E_{\lambda \lambda x}$ are obtained by applying \cref{eq: E_x} to $E_\lambda$ and $ E_{\lambda \lambda}$. Then invariant $N$ can be obtained following \cref{eq: N-3} and \cref{eq: N-4}.

To extract the light effects, ReLU and normalization functions are first applied to filter out minor disturbances. Then, by filtering invariant $N$ with the well-lit mask $M_n$, we obtain the light effects from the well-lit regions. The operations above can be formulated as:
\begin{equation} 
    \begin{aligned}
        & M_{le}  = \mathrm{ReLU}( \frac{N - \mu(N)}{\sigma(N)}) \odot M_n,
    \end{aligned}	
    \label{eq: le}
\end{equation}  
while the well-lit mask are refined: $M_n \leftarrow M_n - M_{le}$.

With the initial disentanglement in \cref{sec: Physical}, we obtain the final disentanglement:  $M_d$, $M_n$, $M_h$ and $M_{le}$. All the masks are stacked to obtain the disentanglement map. By employing the aforementioned techniques and processes, we successfully achieve the disentanglement of various degradation regions, in contrast to naive clustering methods. Our approach, developed based on a physics prior, more closely aligns with real-world scenarios. The visualization results are shown in \cref{fig:dis}.

\subsection{Degradation-Aware Contrastive Learning}
For unpaired image translation, contrastive learning has validated its effectiveness for the preservation of
content. It targets to maximize the mutual information between patches in the same spatial location from the generated image and the source image as below:
\begin{equation} 
\begin{aligned}
     \ell (v, v^{+}, v^{-}) = -\log \frac{\exp (v\cdot v^{+}/\tau )}{\exp (v\cdot v^{+}/\tau ) + \sum_{n=1}^Q \exp (v\cdot v^{-}_n/\tau )} ,
\end{aligned}	
\label{eq: PatchNCE} 
\end{equation}
$v$ is the anchor that denotes the patch from the generated image. The positive example $v^{+}$ corresponds to the source image patch with the same location as the anchor $v$. The negative examples $v^{-}$ represent patches with locations distinct from that of the anchor $v$. $Q$ denotes the total number of negative examples. In our work, the key insight of degradation-aware contrastive learning lies as following: (1) How to sample the anchor, positive, and negative examples. (2) How to manage the focus on negative examples.

\noindent \textbf{Degradation-Aware Sampling.} In this paper, N2D3 selects the anchor, positive, and negative patches under the guidance of the disentanglement results. Initially, based on the disentanglement mask obtained in the \cref{sec: dis}, we compute the patch count for different degradation types, denoting as $K_s, s \in [1,4]$. Then, within each degradation region, the anchors $v$ are randomly selected from the patches of generated daytime images $I_{\mathcal{N}\rightarrow \mathcal{D}}$. The positive examples $v^{+}$ are sampled from the same locations with the anchors in the source nighttime images $I_\mathcal{N}$, and the negative examples $v^{-}$ are randomly selected from other locations of $I_\mathcal{N}$. For each anchor, there is one corresponding positive example and $K_s$ negative examples. Subsequently, the sample set with the same degradation type will be assigned weights and the contrastive loss will be computed in the following steps. 
\begin{figure*}[htbp]
\centering
\includegraphics[width=0.95\linewidth]{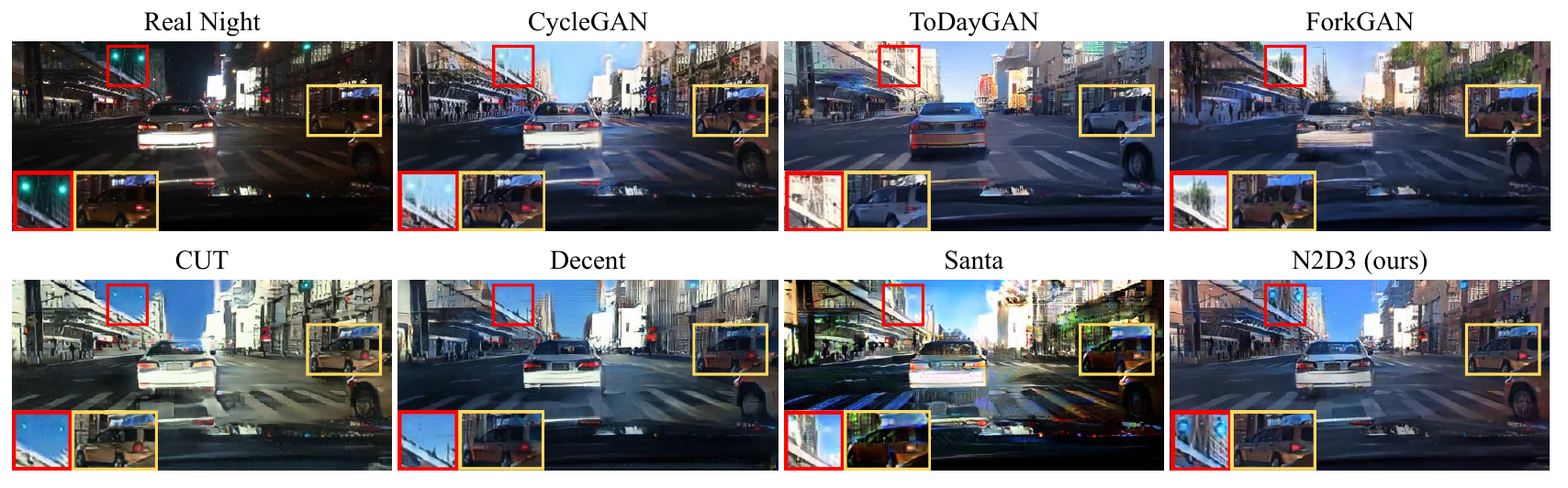} 
\vspace{-0.4cm}
\caption{The qualitative comparison results on the BDD100K dataset.}
\vspace{-0.6cm}
\label{fig: bdd results}
\end{figure*}

\noindent \textbf{Degradation-Aware Reweighting.} Despite the careful selection of anchor, positive, and negative examples, the importance of anchor-negative pairs still differs within the same degradation. A known principle of designing contrastive learning is that the hard anchor-negative pairs (\ie, the pairs with high similarity) should assign higher attention. Thus, weighted contrastive learning can be formulated as:
\begin{flalign}
& \ell (v, v^{+}, v^{-}, w_{n})  \\ 
& = -\log \frac{\exp (v\cdot v^{+}/\tau )}{\exp (v\cdot v^{+}/\tau ) + \sum_{n=1}^Q w_{n} \exp (v\cdot v^{-}_n/\tau )}, 
\nonumber
\label{eq: DegraNCE} 
\end{flalign}
$w_{n}$ denotes the weight of the $n$-th anchor-negative pairs.

The contrastive objective is depicted in the \textit{Similarity Matrix} in \cref{fig:main}.
The patches in different regions are obviously easy examples. We suppress their weights to 0, which transforms the similarity matrix into a blocked diagonal matrix with $diag(A_1, \ldots, A_4)$. Within each degradation matrix $A_s, s\in [1,4]$, a soft reweighting strategy is implemented. Specifically, for each anchor-negative pair, we apply optimal transport to yield an optimal transport plan, serving as a reweighting matrix associated with the disentangled results. It can adaptively optimize and avoid manual design. The reweight matrix for each degradation type is formulated as:
\begin{equation}
    \begin{aligned}
      \min_{w_{ij}, i,j \in [1, K_s]}  [\sum_{i=1}^{K_s} & \sum_{j=1, i\ne j}^{K_s}  w_{ij}\cdot \exp{(v_i \cdot v^{-}_j / \tau)} ], \\
     \sum_{i=1}^{K_s} w_{ij} = 1, &\sum_{j=1}^{K_s} w_{ij}= 1, i, j\in [1, K_s],
    \end{aligned}
\end{equation}

The aforementioned operations transform the contrastive objective to the \textit{Block Diagonal Similarity Matrix} depicted in \cref{fig:main}. As a common practice, our degradation-aware contrastive loss is applied to the $S$ layers of the CNN feature extractor, formulated as:
\begin{equation}
    \begin{aligned}
    \mathcal{L}_{DegNCE}(\mathcal{F}) =  \sum _{l=1}^S  \ell ({v, v^{+}, v^{-}, w_{n}}). 
    \end{aligned}
\end{equation}

\subsection{Other Regularizations}
As a common practice, GANs are employed to bridge the domain gap between daytime and nighttime. The adversarial loss is formulated as:
\begin{equation}
\begin{aligned}
    &\mathcal{L}_{adv}(\mathcal{F}) = ||D(\mathbf{I}_\mathcal{N \rightarrow D})-1||_2^2,\\
    &\mathcal{L}_{adv}(D) = ||D(\mathbf{I}_\mathcal{D})-1||_2^2 + ||D(\mathbf{I}_\mathcal{N \rightarrow D})||_2^2,
\end{aligned}
\end{equation}
where $D$ denotes the discriminator network. The final loss function is formatted as :
\begin{equation}
\begin{aligned}
    \mathcal{L}(\mathcal{F}) &= \mathcal{L}_{adv}(\mathcal{F}) + \mathcal{L}_{DegNCE}(\mathcal{F}), \\
    \mathcal{L}(D) &= \mathcal{L}_{adv}(D) .
\end{aligned}
\end{equation}

%% file: sec/4_exp.tex
\section{Experiments}\label{sec: Experiments}
\subsection{Experimental Settings}

\begin{figure*}[htbp]
\centering
\includegraphics[width=0.95\linewidth]{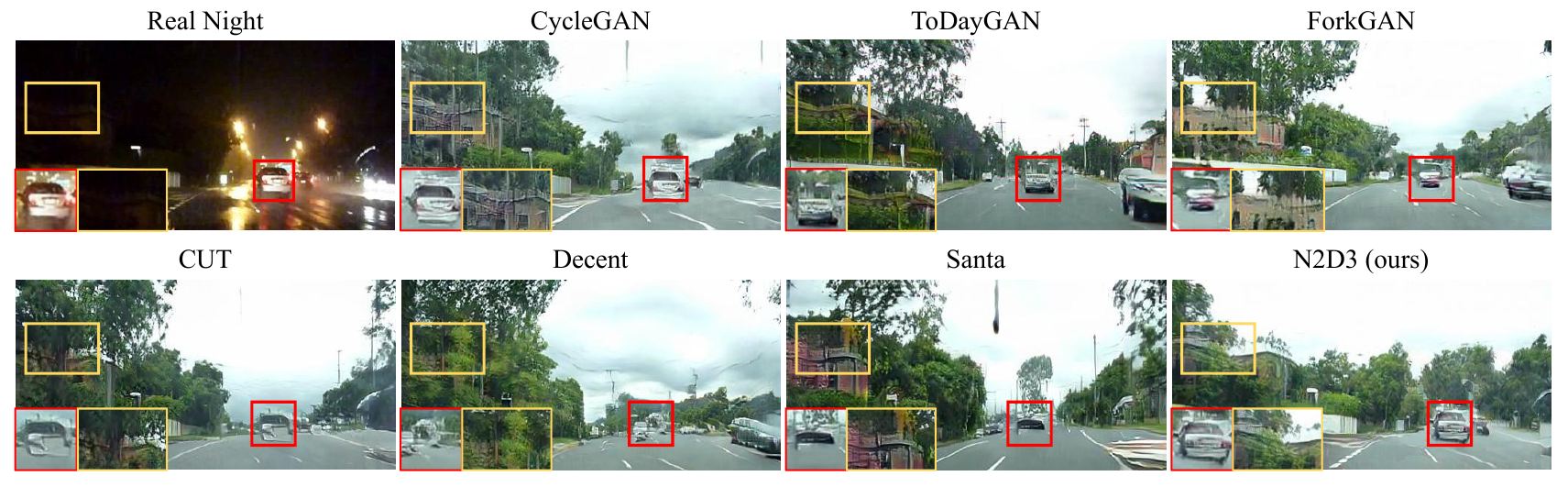} 
\vspace{-0.5cm}
\caption{The qualitative comparison results on the Alderley dataset.}
\vspace{-0.2cm}
\label{fig: alderley results}
\end{figure*}

\begin{table*}[ht]

\centering
\caption{The quantitative results on Alderley and BDD100k. $\downarrow$ means lower result is better.  $\uparrow$ means higher is better.}
\vspace{-0.2cm}
  \resizebox{0.7\textwidth}{!}{
\begin{tabular}{l | c c  c | c c c }
\hline
\toprule \multirow{2}{*}{ \textbf{Methods} } & \multicolumn{3}{c|}{ \textbf{Alderley} } & \multicolumn{3}{c}{ \textbf{BDD100k}} \\ 
&  FID$\downarrow$  & LPIPS$\downarrow$ & SIFT$\uparrow$ & FID$\downarrow$ & LPIPS$\downarrow$ & mIoU$\uparrow$ \\ \hline
Original  & 210 & - & 3.12 & 101 & - & 15.63 \\
\hline
CycleGAN\cite{zhu2017unpaired}(ICCV 17)& 167 & 0.706 & 3.36 & 51.7 & 0.477 & 13.42 \\
StarGAN\cite{choi2018stargan}(CVPR 18) &117 & - & 3.28 & 68.3 & - &-\\
ToDayGAN\cite{anoosheh2019night}(ICRA 19) &104 & 0.770 & 4.14 & 43.8 & 0.577 & 16.77 \\
UGATIT\cite{kim2019u}(ICLR 20) & 170 & - & 2.51 & 72.2 & - & - \\
CUT\cite{park2020contrastive}(ECCV 20) & 64.7 & 0.707 & 6.78 & 55.5 & 0.583 & 9.30  \\
ForkGAN\cite{zheng2020forkgan}(ECCV 20) &61.2 & 0.759 & 12.1 & 37.6 & 0.581 &11.81 \\
AUGAN\cite{kwak2021adverse}(BMVC 21)& 65.2 & - & - & 38.6 & - & - \\
MoNCE\cite{Zhan_2022_CVPR}(CVPR 22)& 72.7 & 0.737 & 6.35 & 40.2 & 0.502 & 17.21 \\
Decent\cite{decent}(NIPS 22)& 76.5 & 0.768 & 6.31 & 40.3 & 0.582 & 10.49\\
Santa\cite{santa}(CVPR 23)&67.1 & 0.757 & 6.93 & 36.9 & 0.559 & 11.03\\
StegoGAN\cite{wu2024stegogan}(CVPR 24)& 82.8 & 0.718 & -  & 89.9 & 0.687& -\\
\hline
Zero-DCE \cite{guo2020zero}(TPAMI 20) & 246.4 & - & 4.34 &90.5 &- &15.90 \\
EnlightenGAN  \cite{jiang2021enlightengan}(TIP 21)& 209.8 & - & 2.00 &103.5 &- &16.10  \\
WCDM \cite{diffll}(ToG 23)& 239.6 & - & 7.10 & 124.3 & - & 16.32  \\
GSAD \cite{hou2024global}(NIPS 23)& 214.7 & - & 6.29 & 116.0 & - & 15.76 \\
\hline
N2D3(Ours) & \bf{50.9} & \bf{0.650} & \bf{16.62} & \bf{31.5} & \bf{0.466} & \bf{21.58}  \\
\hline
\end{tabular}}
\label{tab:alder}
\vspace{-0.6cm}
\end{table*}
\textbf{Datasets.} Experiments are conducted on the two public datasets BDD100K \cite{yu2020bdd100k} and Alderley \cite{alderley}. \textbf{Alderley} dataset consists of images captured along the same route twice: once on a sunny day and another time during a stormy rainy night. The nighttime images in this dataset are often blurry due to the rainy conditions, which makes Night2Day challenging. 

\noindent\textbf{BDD100K} dataset is a large-scale high-resolution autonomous driving dataset. It comprises 100,000 video clips under various conditions. For each video, a keyframe is selected and meticulously annotated with details. We reorganized this dataset based on its annotations, resulting in 27,971 night images for training and 3,929 night images for evaluation. 

\noindent\textbf{Evaluation Metric.} Following common practice, we utilize the \emph{Fréchet Inception Distance} (FID) scores \cite{heusel2017gans} to assess whether the generated images align with the target distribution. This assessment helps determine if a model effectively transforms images from the night domain to the day domain. Additionally, we seek to understand the extent to which the generated daytime images maintain structural consistency compared to the original inputs. To measure this, we employ SIFT scores, 
mIoU scores and LPIPS distance \cite{LPIPS}. 

\noindent\textbf{DownStream Vision Task.}
Two downstream tasks are conducted. In the Alderley dataset, GPS annotations indicate the locations of two images, one in the nighttime and the other in the daytime, as the same. We calculate the number of SIFT-detected key points between the generated daytime images and their corresponding daytime images to measure if the two images represent the same location. The BDD100K dataset includes 329 night images with semantic annotations. We employ Deeplabv3 pretrained on the Cityscapes dataset as the semantic segmentation model \cite{chen2017rethinking}, then perform inference on our generated daytime images without any additional training and compute the mIoU (mean Intersection over Union).


\subsection{Results on Alderley}
We first apply Night2Day on the Alderley dataset, a challenging collection of nighttime images captured on rainy nights. In Figure \ref{fig: alderley results}, we present a visual comparison of the results. CycleGAN \cite{zhu2017unpaired} and CUT \cite{park2020contrastive} manage to preserve the general structural information of the entire image but often lose many fine details. ToDayGAN \cite{anoosheh2019night}, ForkGAN \cite{zheng2020forkgan}, Decent \cite{decent}, and Santa \cite{santa} tend to miss important elements such as cars in their results.

In \cref{tab:alder}, translation methods and enhancement methods are compared, considering both visual effects and keypoint matching metrics. Our method showcases \textbf{an improvement of 10.3 in FID scores and 4.52 in SIFT scores} compared to the previous state-of-the-art. This suggests that N2D3 successfully achieves photorealistic daytime image generation, underscoring its potential for robotic localization applications. The qualitative comparison results are demonstrated in \cref{fig: alderley results}. N2D3 excels in generating photorealistic daytime images while effectively preserving structures, even in challenging scenarios such as rainy nights in the Alderley.


\begin{figure*}[ht]
\centering
\includegraphics[width=0.99\linewidth]{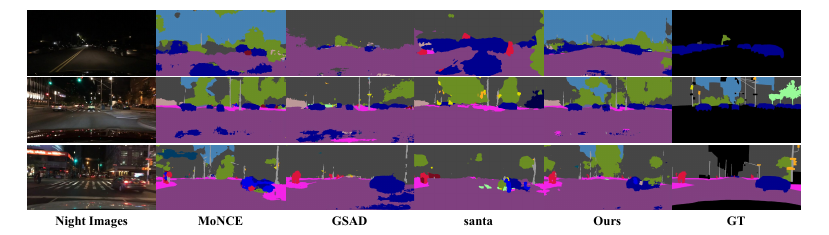} 
\vspace{-0.2cm}
\caption{The visual comparison of nighttime segmentation. The results demonstrate that our method learns a more accurate daytime distribution, providing the greatest performance improvement over the typical model trained on daytime images and yielding superior segmentation results.}
\vspace{-0.5cm}
\label{fig:seg_results}
\end{figure*}

\subsection{Results on BDD100K}

We conducted experiments on a larger-scale dataset, BDD100K, focusing on more general night scenes. The qualitative results can be found in \cref{fig: bdd results}. CycleGAN, ToDayGAN, and CUT succeed in preserving the structure in well-lit regions. ForkGAN, Santa, and Decent demonstrate poor performance in such challenging scenes. Regretfully, none of them excel in handling light effects and exhibit weak performance in maintaining global structures. With a customized design specifically addressing light effects, our method successfully preserves the structure in all regions. 

The quantitative results are presented in \cref{tab:alder}. As the scale of the dataset increases, all the compared methods show an improvement in their performance. Notably, N2D3 demonstrates the best performance with \textbf{a significant improvement of 5.4 in FID scores}, showcasing its ability to handle a broader range of nighttime scenes.

We also investigate the potential of Night2Day in enhancing downstream vision tasks in nighttime environments using the BDD100K dataset. The quantitative results are summarized in \cref{tab:alder}. The enhancement methods demonstrate a slight improvement in segmentation results, while some image-to-image translation methods have a negative impact on performance. N2D3 exhibits the best performance in enhancing nighttime semantic segmentation with \textbf{a remarkable improvement of 5.95 in mIoU} compared to inferring the segmentation model directly on nighttime images. The visualization results are shown in \cref{fig:seg_results}, highlighting its benefits for downstream tasks and its potential for wide-ranging applications.


\begin{table}[t]
\centering
\caption{The quantitative results of ablation on the main component of degradation-aware contrastive learning. (a) denotes the degradation-aware sampling, and (b) denotes the degradation-aware reweighting.}
\vspace{-0.2cm}
\resizebox{0.95\linewidth}{!}{
\begin{tabular}{cc|cc|ccc}
\hline
\multicolumn{2}{c|}{ Main Component}      & \multicolumn{2}{c|}{BDD100K} & \multicolumn{3}{c}{Alderley} \\ \hline
  (a) & (b) & FID   & LPIPS  & FID     & LPIPS    & SIFT    \\ \hline
 \XSolidBrush   &\XSolidBrush     &55.5 &0.583      & 64.7    & 0.707    &  6.78      \\
 \Checkmark   &\XSolidBrush & 36.9  & 0.495 &  56.6  & 0.698     &   16.52     \\
\Checkmark   & \Checkmark & 31.5  & 0.466  &50.9    & 0.650         &  16.62      \\ \hline
\end{tabular}}
\label{tab:abaltion on degnce}
\vspace{-0.6cm}
\end{table}

\subsection{Ablation Study}

We present additional ablation studies on the four components, as detailed in \cref{tab:abaltion on degnce} and \cref{tab:abaltion on invariant}. The studies reveal that while performance slightly improves with refined classification into four clusters, a more accurate segmentation based on our physical model significantly enhances performance and achieves optimal results. The challenge arises from the similarity in intensity between light effect regions and well-lit areas, making it difficult to differentiate them using a simple k-Means. Our physical prior, which extracts features beyond intensity, enables better subdivision and contributes significantly to the final performance.

\noindent\textbf{Ablation on the main component of degradation-aware contrastive learning.} 
The core design of the degradation-aware contrastive learning module relies on two main components: (a) degradation-aware sampling, and (b) degradation-aware reweighting. As shown in \cref{tab:abaltion on degnce}, when degradation-aware sampling is exclusively activated, there is a noticeable decrease in FID on both datasets compared to the baseline (no components activated). Notably, the combination of degradation-aware sampling and reweighting achieves the lowest FID on both BDD100K and Alderley, indicating the effectiveness of degradation-aware sampling in conjunction with degradation-aware reweighting.


\begin{table}[t]
\centering
\caption{The quantitative results of ablation on the physical prior invariant.  $L$ denotes illuminance map and $N$ denotes the physical prior invariant.}
\vspace{-0.2cm}
\resizebox{0.95\linewidth}{!}{
\begin{tabular}{cc|cc|ccc}
\hline
\multicolumn{2}{c|}{ Invariant Type}      & \multicolumn{2}{c|}{BDD100K} & \multicolumn{3}{c}{Alderley} \\ \hline
 L & N  & FID   & LPIPS   & FID     & LPIPS    & SIFT    \\ \hline
\XSolidBrush &  \XSolidBrush  &55.5   & 0.583   & 64.7    & 0.707    &  6.78      \\
\Checkmark &  \XSolidBrush   & 49.1  & 0.592  & 62.9    & 0.726    &  9.83      \\
\Checkmark & \Checkmark    & 31.5  & 0.466  &50.9    & 0.650         &  16.62      \\ 
\hline
\end{tabular}}
\label{tab:abaltion on invariant}
\vspace{-0.6cm}
\end{table}
\noindent\textbf{Ablation on the type of the invariant in disentanglement.} To explore different invariants for obtaining degradation-disentangled prototypes, we conduct an ablation study on the type of invariant. As shown in \cref{tab:abaltion on invariant}, when $L$ is enabled, the FID decreases from 55.5 to 49.1 on BDD100K and from 64.7 to 62.9 on Alderley. This suggests that incorporating illuminance maps helps in reducing the perceptual gap between generated and source nighttime images. When $N$ is activated, there is a consistent improvement in FID on both datasets, indicating that considering physical priors invariant contributes to more realistic image generation. The combination of both illuminance map and physical prior invariant results in the lowest FID on both datasets, showcasing the complementary nature of these degradation types in improving contrastive learning.

%% file: sec/5_conclusion.tex
\section{Conclusion}
This paper introduces a novel solution for the Night2Day image translation task, focusing on translating nighttime images to their corresponding daytime counterparts while preserving semantic consistency. To achieve this objective, the proposed method begins by disentangling the degradation presented in nighttime images, which is the key insight of our method. To achieve this, we contribute a degradation disentanglement module and a degradation-aware contrastive learning module. Our method outperforms the existing state-of-the-art, which shows the effectiveness and the superiority of the insight to disentangle the degradation. 